\documentclass{article}

\usepackage{arxiv}
\usepackage[numbers]{natbib}

\usepackage{graphicx}
\usepackage{booktabs}
\usepackage{hyperref}
\hypersetup{colorlinks=true, linkcolor=black, citecolor=blue, urlcolor=blue}
\usepackage{url}
\usepackage{amsfonts}
\usepackage{amsmath}
\usepackage{amssymb}
\usepackage{microtype}
\usepackage{xcolor}
\usepackage{multirow}
\usepackage{subcaption}
\usepackage{float}
\usepackage{tikz}
\usetikzlibrary{arrows.meta,positioning,fit,calc,backgrounds}
\usepackage{algorithm}
\usepackage{algpseudocode}

\graphicspath{{figures/}}

\title{PosterHarness: Turning Scientific Poster Generation into an Auditable Instruction-Following Benchmark}
\renewcommand{\shorttitle}{PosterHarness}
\renewcommand{\headeright}{}

\author{Zijian Wang\thanks{\texttt{wangzijian@stu.pku.edu.cn}}\quad
Dawei Fu\quad Youpeng Wu\quad Zixun Kou\\
Linrui Chen\quad Ruobing Jiang\quad Tianyi Yang\thanks{\texttt{tyyang99@pku.edu.cn}}\quad
Qiang Li\thanks{\texttt{qliphy@gmail.com}}\\
School of Physics, Peking University}

\begin{document}

\maketitle

\begin{abstract}
Text-rich image models can now design poster-scale layouts, but we lack ways to \emph{measure}
whether they honor the contracts scientific communication demands: rendering legible labels,
following prescribed figure aspect ratios, and---above all---abstaining from drawing fabricated
scientific figures. We present \textsc{PosterHarness}, an auditable harness that reframes
scientific poster generation as a battery of measurable instruction-following tasks for
text-rich image models, together with a pilot benchmark and a failure taxonomy. Our target
is not to claim a production poster-authoring system, but to provide a reusable way to
audit where these models succeed and fail on scientific layout contracts.

\textsc{PosterHarness} uses a \textbf{placeholder-first} contract to separate two jobs
that image models otherwise conflate. The image model performs visual-summary design:
typography, reading path, color, and background. It does \emph{not} draw data-bearing
scientific figures. Instead, every figure region must be an empty labeled placeholder, and
a deterministic compositor later inserts real source-paper figures at the detected
coordinates. This makes specific properties measurable---placeholder count and ID accuracy,
blankness, aspect-ratio compliance, abstention from model-synthesized scientific graphics,
public-text hygiene, and source-figure provenance of accepted regions, conditional on
correct extraction, assignment, and detection. Failures are logged as explicit rejections
rather than hidden inside plausible-looking output.

This formulation also gives a practical visual-summary use case: a long paper can be
turned into a one-page visual overview whose scientific figure regions remain traceable to
the source document.

We instantiate the harness on a 12-paper set (6 HEP analyses and 6 non-HEP AI/ML-adjacent
papers) and report three kinds of evidence. (i) A small counterfactual prompt probe shows
that the placeholder contract drives synthesized scientific figures counted by a
vision-language model (VLM) from 34 to 0 across three papers. (ii) A run-level failure
taxonomy parsed from saved manifests records which contracts actually block acceptance:
placeholder geometry, placeholder quality assurance (QA), template critic, and public text.
(iii) A comparison
with Paper2Poster shows a trade-off: PosterHarness produces higher-resolution artifacts
with lower white-canvas fraction and stronger single-VLM visual preference, while the
deterministic baseline retains slightly more PosterQuiz-style information and runs faster.
We report this as a regime characterization, not a superiority claim. We release all
artifacts, prompts, manifests, and audit scripts as a reusable evaluation component for both
poster-style visual summarization and model compliance analysis.
\end{abstract}

\section{Introduction}

A conference poster faces a structural tension. It has one canvas but must
carry the entire core of a 10-to-30-page paper---data, figures, conclusions.
At the same time it must hold enough visual presence to compete for attention
in a dense poster hall. These two demands---information density and visual
quality---pull in opposite directions, making poster design a time-consuming
task that requires substantial design expertise.

Automating this process requires more than unconstrained image synthesis. A usable
poster-generation system must follow a long, multi-part design brief rather than a
single short prompt; render titles, section headers, bullets, captions, and
numerical badges legibly; place multiple figure slots at distinct prescribed aspect
ratios; and respect the visual conventions of different disciplines. Standard
text-to-image benchmarks test few of these requirements, and---crucially---there is
no standard way to \emph{measure} whether a model satisfies them on scientific
content. That measurement gap, rather than poster authoring as a product category,
is what this paper sets out to address.

Prior work falls broadly into two families, each with different limitations.

The first was deterministic rendering, best represented by
Paper2Poster~\cite{pang2025paper2poster}. It borrowed multi-agent workflow ideas
from LLM systems~\cite{wu2023autogen, hong2023metagpt, li2023camel} to
decompose the task: one module parses the paper, another plans the layout, a
third renders each panel in PPTX. The advantage is reliability: text placement
is handled by a layout engine, and figures are copied from source files rather than
synthesized by a generative image model. The cost is the medium itself. PPTX enforces a slide-grid
aesthetic---uniform blocks, rigid multi-column layouts, a narrow color
palette---and the result often resembles a slide deck rather than an editorial design.
Paper2Poster's own user study identified ``reader engagement as the primary
aesthetic bottleneck''~\cite{pang2025paper2poster}.

The second approach---image generation models---took the opposite path. They
could produce visually striking outputs, but they could not render readable text
at poster scale~\cite{chen2023textdiffuser, tuo2023anytext, yang2023glyphcontrol, chen2024textdiffuser2, shimoda2024typer}.
Worse, they could not be trusted to draw scientific figures. When a diffusion
model is asked to produce a distribution plot, a scatter plot, or an exclusion
limit curve, it can produce an artifact that is visually plausible but numerically
unsupported. The curve follows the statistics of its training data, not a
measurement, and the error bars have no source in the paper.
Third-party evaluations of GPT-4o's native image generation later confirmed
that even the strongest models are unreliable on knowledge-intensive scientific
illustrations~\cite{cao2025gpt4o}. In scientific communication there is no room
for ``close enough'': a figure either comes from real data or it is wrong.

Recent OpenAI image-generation systems changed the feasible design space. GPT-4o's
native image generation in March 2025~\cite{openai2025imagegen} was an important
step, but its text rendering and long-instruction following remained unreliable for
production use~\cite{cao2025gpt4o}. ChatGPT Images 2.0 (image-2.0) in April
2026~\cite{openai2026chatgptimages2systemcard}, together with the GPT Image
API~\cite{openai2026imagegenerationdocs}, materially improved the ability to render
dense poster text and follow multi-constraint design briefs. The \textsc{BizGenEval}
benchmark evaluated 26 systems on slides, posters, and scientific graphics and
confirmed this judgment~\cite{li2026bizgeneval}. Critically, BizGenEval's finding
is not that one particular model underperforms: it reports ``substantial
capability gaps'' across \emph{all} evaluated systems on structured
multi-constraint design tasks. This means scientific-figure trustworthiness is
not merely a model-capacity problem; it is a systematic architectural gap. These developments make image-model-based poster layout
credible, but they do not make model-synthesized scientific figures acceptable.

This gap motivates the central principle of our work: separate two jobs at the
architectural level, not through a more detailed prompt. The first job is
\emph{visual-summary design}: organizing the paper's narrative, typography, color, and
reading path on a poster-scale canvas. The second job is \emph{scientific figure
grounding}: placing data-bearing figures whose pixels must come from the source paper. The
separation is useful only if it also makes the model's compliance \emph{measurable}.

\textsc{PosterHarness} is built on a \textbf{placeholder-first} contract. The
image model handles tasks for which it is comparatively well suited---global composition,
color, typography, background art, editorial style---subject to one hard prohibition: every region
destined to carry a scientific figure must appear as an empty placeholder box.
The box contains exactly three things: a figure ID (e.g., \texttt{[FIG 01]}), a
short public label, and the target aspect ratio. No curves, axes, histograms,
Feynman lines, microscopy images, or any data-bearing content is permitted
inside those slots. After generation, a vision-language-model (VLM)-based template critic scores the
template on overall quality, artistry, information density, and
placeholder-contract compliance; failures trigger regeneration with structured
feedback. Once a template passes, a deterministic replacement engine detects the
exact pixel coordinates of each placeholder box and overlays the real
source-paper figures at those coordinates---aspect ratio locked, position locked,
no stretching and no cropping. This pipeline enforces three narrower, auditable
properties:
(i) every accepted figure region on the poster contains a real paper asset,
not a model-synthesized plot; (ii) the generative component is limited to creative visual design, while
correctness-sensitive work stays on the deterministic path; (iii) any placement failure---wrong aspect ratio, non-blank
slot, containment violation---is caught and rejected before compositing, never
hidden inside a plausible-looking result.

Framed this way, \textsc{PosterHarness} is not primarily a poster-authoring product,
but an \emph{auditable benchmark} of scientific layout instruction-following: each property above is a
condition that a text-rich image model either satisfies or visibly fails, and the harness logs every
such success and rejection. The poster is the vehicle; the contribution is a reusable way to
measure where these models break the contracts scientific communication requires.

The framework also handles a practical engineering issue: a single set of
layout rules cannot simultaneously serve high-energy physics papers and computer
science papers. We introduce declarative \textbf{domain profiles}---HEP, CS/ML,
biology, astrophysics, mathematics, chemistry, plus a generic fallback. Each
profile is a YAML configuration that specifies four things: the poster's
narrative spine (HEP: motivation $\to$ strategy $\to$ result $\to$
interpretation; CS/ML: problem $\to$ method $\to$ experiments $\to$
limitations); the hierarchy of figure placement; what expert readers in that
field most need to learn from the poster; and what counts as forbidden
decoration (no fake Feynman graphs in a physics poster, no invented microscopy
panels in a biology poster). Profiles are data, not code---adding a new
discipline does not touch the pipeline. The spines are manually specified defaults based on
common paper structures in each field, not an empirical claim that all CS/ML or HEP papers
follow one universal story. On top of profiles, \textbf{density presets} (standard and
\texttt{hep\_dense}) control how much paper-specific detail survives into the final poster,
allowing a HEP analysis to carry signal-region definitions and systematic-uncertainty tables
while a CS paper emphasizes architecture diagrams.

The final line of defense is a \textbf{multi-stage quality-assurance (QA) chain}:
template critic with regeneration, placeholder geometry verification, containment audit
after figure insertion, and public-text review before export. Every run saves its prompts,
structured specs, figure-assignment decisions, QA reports, and a status manifest---so that an
accepted poster can be traced through the failures and retries that led to acceptance. The
public repository\footnote{\url{https://github.com/tyy99phy/paper_poster_harness}} records these prompts, manifests, and run-level traces so that the textual
corpus can be inspected without re-running the expensive image-generation stages.

Our contributions are:

\begin{enumerate}
  \item \textbf{A measurable instruction-following formulation of scientific poster
    generation.} We turn an open-ended generation task into a set of auditable contracts---placeholder
    count and ID accuracy, blankness, aspect-ratio compliance, abstention from
    model-synthesized scientific graphics, public-text hygiene, and source-figure
    provenance of accepted regions---each checked by deterministic geometry tests or VLM
    audit, with failures surfaced as explicit, categorizable rejections rather than hidden
    inside plausible output. This makes ``did the image model follow the scientific layout
    contract?'' a measurable question.
  \item \textbf{The placeholder-first mechanism that makes those contracts enforceable.} The
    image model designs typography, reading path, and background but leaves every scientific
    figure region as a labeled blank placeholder; a deterministic compositor inserts real
    source-paper figures at detected coordinates. A small three-paper probe isolates the
    mechanism's effect, driving synthesized scientific figures counted by a VLM from 34 to 0.
    Declarative domain profiles and density presets extend the contracts to new fields without
    code changes.
  \item \textbf{A released pilot benchmark, failure taxonomy, and audit artifacts.} We
    instantiate the harness on a 12-paper set (6 HEP analyses, 6 non-HEP AI/ML-adjacent
    papers) and release paired posters, prompts, run manifests, a manifest-derived failure
    taxonomy, and low-cost audit scripts as a reusable evaluation component. The artifacts also
    demonstrate a source-grounded visual-summary use case: a paper can be summarized as a
    one-page, poster-style visual overview without asking the image model to invent scientific
    evidence. A comparison with Paper2Poster~\cite{pang2025paper2poster} is reported as a
    trade-off, not a superiority claim.
\end{enumerate}

\section{Related Work}

Scientific poster generation sits at the intersection of three research lines:
automatic poster-generation systems, controllable image generation and scientific-figure
hallucination, and domain adaptation for scientific content. We cover each briefly, focusing
on their relationship to our contribution rather than re-summarizing what the Introduction
already described.

\subsection{Automatic Poster Generation}

The most direct reference is \textsc{Paper2Poster}~\cite{pang2025paper2poster},
which brought multi-agent workflow ideas~\cite{wu2023autogen, hong2023metagpt,
li2023camel} to poster generation through a parser-planner-renderer decomposition.
Its deterministic PPTX backend provides strong control over text placement and figure provenance but is
constrained by the medium's visual ceiling. Our work shares Paper2Poster's
structured-pipeline philosophy but diverges at two points: (i) we replace the
PPTX renderer with an image model to gain visual freedom, and (ii) we insert a
placeholder contract and deterministic replacement between generation and
compositing to protect scientific figure integrity.

A parallel and largely concurrent line of work targets \emph{graphic-design} posters
rather than scientific provenance. DreamPoster~\cite{hu2025dreamposter} casts
image-conditioned poster design as a unified generative problem, and
EfficientPosterGen~\cite{tang2026efficientpostergen} pursues semantic-aware,
lower-cost generation. On the measurement side,
PosterIQ~\cite{feng2026posteriq} contributes a design-perspective benchmark for poster
understanding and generation, and PosterReward~\cite{lai2026posterreward} trains a
dedicated reward model for high-quality graphic-design output. These systems optimize
and score aesthetic and layout quality for commercial or general-purpose posters, where
any pictorial element may be freely synthesized. Our setting differs in kind: the
scientific figures on the poster must be \emph{real} source assets, so a freely
generative or reward-optimized design pipeline does not directly apply. We nonetheless
view learned design-quality evaluators such as PosterReward as a promising replacement
for the single general-purpose VLM judge used in our pilot, because a task-specific
evaluator is less likely to share generation biases with the image model under test.

\subsection{Controllable Generation and Figure Hallucination}

The central assumption behind placeholder-first---that image models should not
be trusted to render scientific figures---is supported by converging evidence.
On the control side, ControlNet~\cite{zhang2023controlnet} and
Prompt-to-Prompt~\cite{hertz2022prompt2prompt} achieve structural constraints
on generated images through spatial conditioning and cross-attention
manipulation, respectively. But these methods address \emph{where} to draw in
natural images, not \emph{whether} to draw in scientific ones. On the evaluation
side, Cao et al.~\cite{cao2025gpt4o} document the frequency of factual errors
produced by native image models on scientific illustrations and mathematical
plots. The LLM-as-Judge paradigm~\cite{zheng2023judge} further suggests that
VLMs can serve as low-cost quality proxies for some assessment tasks, which
directly informs our design of the template critic and QA pipeline.

\subsection{Domain Adaptation}

The benefits of domain adaptation are well established for text---SciBERT~\cite{beltagy2019scibert}
improves NER, classification, and parsing across scientific fields through
continued pretraining. In the visual domain, however, adaptation work remains
focused on document parsing rather than generation: PubLayNet~\cite{zhong2019publaynet}
provides large-scale layout-annotated data but does not involve generative
models. Our domain-profile mechanism attempts to transplant the SciBERT
philosophy of ``encode domain-specific knowledge'' into image-generation
prompt design---not by training a domain-specific model, but by making a
single model operate under different declarative rulesets per discipline.

\section{Method}

\subsection{Problem Formulation}

Given a scientific paper $P$ as input (arXiv ID, local PDF, or PPTX), the goal is to
produce a poster-style visual summary $H$ that satisfies three requirements:

\begin{itemize}
  \item \textbf{Source-figure provenance} ($R_1$): All accepted scientific figure regions
    in $H$ should be populated by real source assets from $P$, not model-generated plots.
  \item \textbf{Layout stability and readability} ($R_2$): Text should not overflow, overlap
    figures, or become illegible; the poster should follow a clear reading path.
  \item \textbf{Domain-profile extensibility} ($R_3$): The system should adapt visual grammar
    and content priorities through declarative domain profiles rather than paper-specific code.
\end{itemize}

We use the term source-figure provenance deliberately. It does not imply full poster-level
factuality: visible text, numerical summaries, conclusions, and figure-caption alignment can
still be wrong if LLM planning, source extraction, or figure assignment fails. Those risks are
handled through QA and audit stages, but they are not formal guarantees.

\subsection{Overall Framework}

\begin{figure}[t]
\centering
\includegraphics[width=\textwidth]{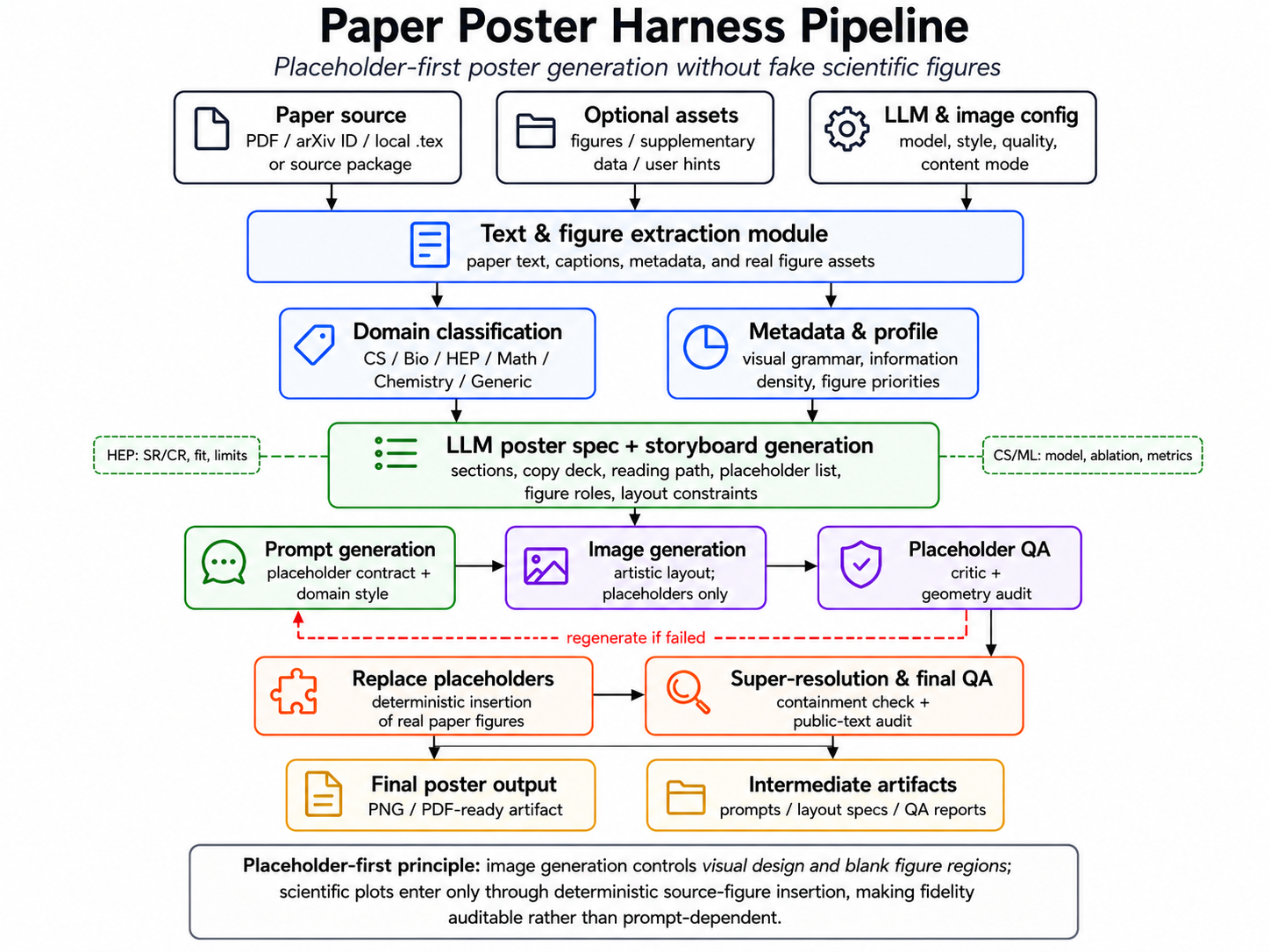}
\caption{Overview of the \textsc{PosterHarness} pipeline. Inputs are first converted into text and figure assets, then routed through domain-aware LLM planning. The upper half produces and audits a placeholder template: the image model may design the visual layout but must leave labeled blank figure slots. The lower half consumes the detected slot geometry: deterministic insertion places real source figures, followed by containment QA, final QA, and export.}
\label{fig:pipeline}
\end{figure}

The \textsc{PosterHarness} pipeline (Figure~\ref{fig:pipeline}) consists of the following stages.

\paragraph{Stage 1: Paper ingestion.} The input paper is downloaded from arXiv or loaded
locally. Text is extracted with PyMuPDF (PDF) or python-pptx (PPTX). Figure assets are
collected automatically from source directories, rendered PDF page images, and embedded PDF
images, filtered to exclude images with width or height below 96~px, capped at 48 assets per
paper. For arXiv papers, the LaTeX source bundle is also downloaded to locate the main
\verb|.tex| file and figure directories.

\paragraph{Stage 2: Domain classification.} An LLM classifier receives a structurally sampled
excerpt of the paper text (title, key sentences, figure captions, truncated to 12,000
characters) and the asset manifest. It outputs a domain profile---\texttt{hep},
\texttt{cs\_ml}, \texttt{bio}, \texttt{astro}, \texttt{math}, \texttt{chemistry}, or
\texttt{generic}---with a confidence score and classification rationale. Each profile is a
YAML-defined ruleset covering four categories: visual grammar (the poster's narrative spine),
figure composition rules (placeholder hierarchy and multi-panel treatment), text priorities
(what experts in the field expect to see), and decorative constraints (what the model may not
draw as decoration). The classifier is conservative: papers it cannot reliably place default
to \texttt{generic}.

\paragraph{Stage 3: Planning artifacts.} The LLM produces seven structured planning documents
in sequence, each constrained to a JSON Schema through the model's structured-output mode. Key documents include:
the \textit{poster spec}---complete layout description with sections, placeholders, and text,
where each placeholder carries an \texttt{id} (e.g., \texttt{"FIG 01"}), a \texttt{section}
index, a public \texttt{label}, and a target \texttt{aspect} (e.g., \texttt{"2.5:1 wide"} or
\texttt{"1:1 square"});
the \textit{storyboard}---narrative spine, \texttt{hero\_section} designation, reading order
in the \texttt{layout\_tree}, per-section \texttt{text\_budget}, and visual role assignment;
the \textit{copy deck}---authoritative visible-text units, each carrying a \texttt{type}
(e.g., \texttt{bullet}, \texttt{badge}, \texttt{hero\_headline}), a \texttt{priority} tier
(\texttt{must}, \texttt{should}, or \texttt{could}), and a \texttt{max\_chars} constraint;
and the \textit{figure selection}---mapping from \texttt{placeholder\_id} to a source
\texttt{asset} path with its measured \texttt{aspect}. In the released configuration, this
selection is capped at eight placeholders (\texttt{autoposter.max\_figures=8}). We treat this
as a layout-stability budget rather than a scientific-content optimum: pilot runs indicate
that asking the image model to place too many independent figure regions on one poster canvas
reduces adherence to per-placeholder aspect ratios, blankness, and ID constraints, especially
when wide, square, and portrait slots are mixed. Under the \texttt{hep\_dense} density preset,
an additional \textit{content outline} is enabled (up to 30 paper-specific facts and 6 key
formulas), and copy-deck capacity is raised from 34 to 42 units.

\paragraph{Stage 4: Image generation.} The assembled prompt has four layers: a base design
brief (overall poster style, typography, color system), domain visual grammar (narrative
spine and composition rules), the placeholder contract (detailed below), and concrete content
constraints extracted from the storyboard and copy deck. The model is asked to produce a
1024$\times$1536 template, with 2 candidate variants per generation. The placeholder contract
includes explicit prohibitions:
\begin{quote}
``Do NOT draw, approximate, summarize, miniaturize, stylize, or recreate ANY scientific
figure content.'' \\
``Every location where a plot/diagram/table/image should go must be a clean rectangular
placeholder panel.'' \\
``Each placeholder panel must contain ONLY three text elements: the exact ID such as
[FIG 01], the intended content label, and the requested aspect ratio.''
\end{quote}
The prompt also carries per-placeholder geometry constraints: ``A box labeled 1:1 must be
visibly square; a box labeled 2.5:1 must be about 2.5 times wider than tall,'' with explicit
counter-examples to avoid: ``Rejected geometry examples: a 2.5:1 plot as a 950$\times$90
ribbon, a 1:1 result as a 300$\times$190 landscape card.'' All figure-containing cards use
light neutral surfaces (warm white, pearl, or very pale blue); dark colors and saturated fills
are permitted only in outer frames, side rails, and header accents.

\paragraph{Stage 5: Template critic and regeneration.} Each generated template is scored by a
VLM-based critic on four dimensions: overall quality (threshold 0.72), artistry (0.65),
information density (0.65), and placeholder-contract compliance (0.75). A template failing
any dimension receives structured critique, which is appended to the prompt for regeneration.
This cycle repeats up to 2 rounds before the pipeline reports failure.

\paragraph{Stage 6: Placeholder detection and geometry verification.} An LLM detects
pixel-level bounding boxes for each placeholder, reporting a confidence score per detection
(minimum 0.15). A deterministic geometry check evaluates IoU and center-distance between each
detected box and its expected location; a detection is rejected when IoU falls below 0.06 and
center-distance fraction exceeds 0.22. Aspect-ratio verification allows 20\% tolerance and
explicitly checks that square placeholders (ratio 0.9--1.1) are not stretched into banners.
Inter-placeholder gutter spacing is also verified to prevent adjacent figures from intruding
into each other after insertion.

\begin{figure}[!htbp]
\centering
\includegraphics[width=\textwidth]{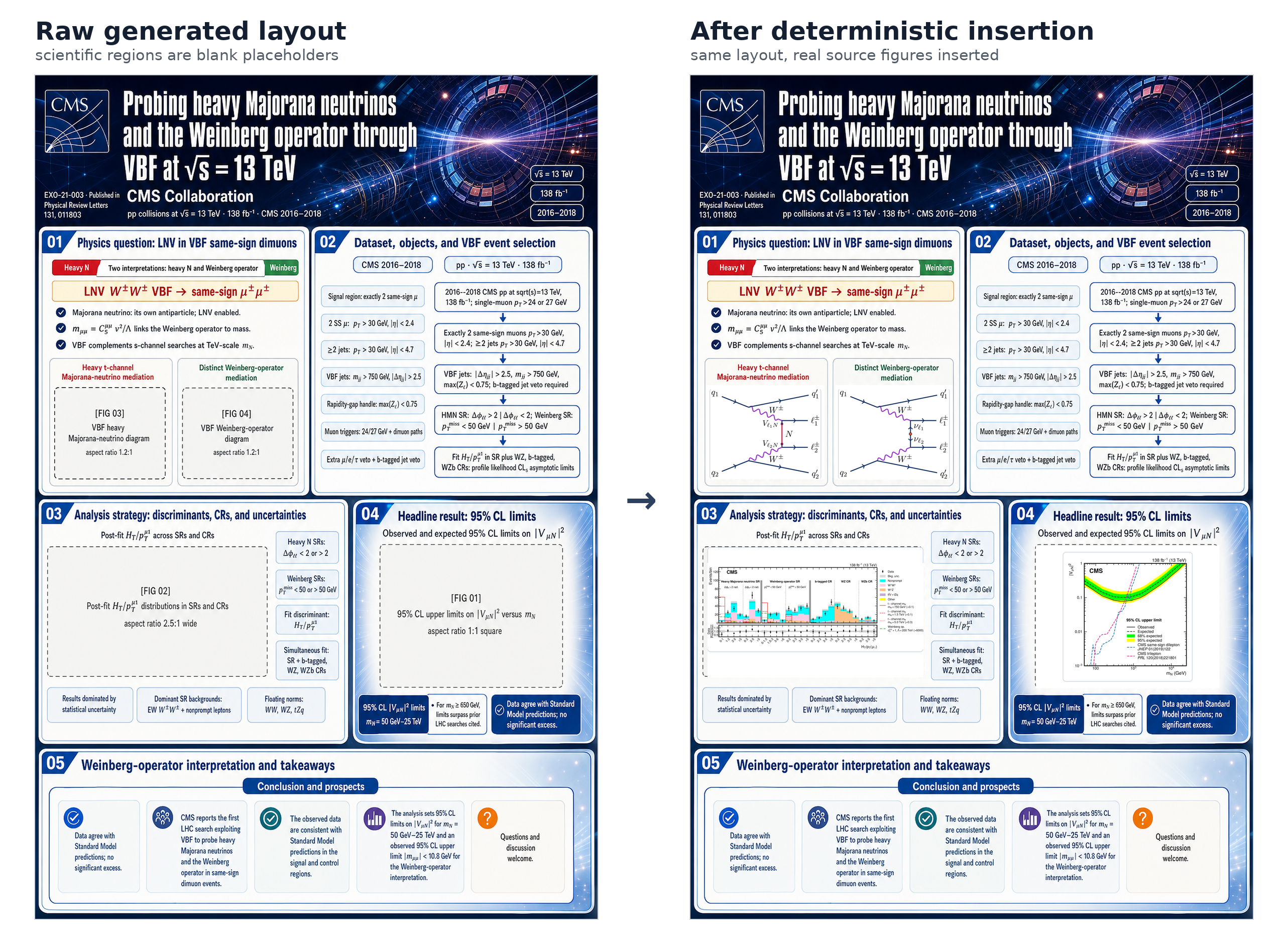}
\caption{Placeholder template (left) and final poster after source-figure insertion (right).}
\label{fig:layout_replacement}
\end{figure}

\paragraph{Stage 7: Deterministic figure replacement.} The replacement engine operates in
three steps per placeholder. First, it computes four bounding boxes: the target placement box,
an erasure cleanup box, an optional decorative outer frame, and the final visual frame.
Second, each target box is validated to be strictly contained within its detected placeholder boundary
(with no inward margin); violations raise hard errors and reject the candidate.
Third, source figures are resized preserving aspect ratio, pasted at target coordinates, and
optionally blended with soft borders and shadows. For square result placeholders, an
artifact-only eraser is used when the cleanup area exceeds 1.85$\times$ the frame area, to
avoid an unnatural white slab in the generated art background. For high-resolution output
(4096$\times$6144), the template is first upscaled 4$\times$, then source figures are pasted
at the scaled coordinates to preserve figure sharpness. Figure~\ref{fig:layout_replacement}
illustrates the placeholder-to-final-poster transition.

\begin{figure}[!htbp]
\centering
\includegraphics[width=\textwidth]{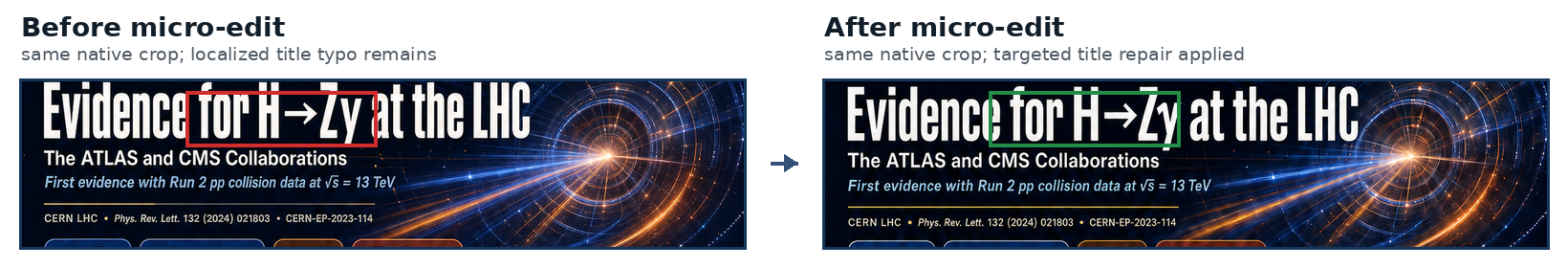}
\caption{Micro-repair example: a localized header defect is corrected via targeted image edit.}
\label{fig:micro_repair}
\end{figure}

\paragraph{Stage 8: Final QA and micro-repair.} Final QA checks three aspects: public-text
compliance (against an 18-phrase default blocklist including ``TODO,'' ``replace placeholder,''
and ``to be replaced,'' which domain profiles may extend), visual coherence, and figure
integration quality. Localized defects such as title typos or minor color bleed trigger the
micro-repair module, which supports three operation types: \texttt{text\_patch} (overlay text
at a target region), \texttt{text\_box} (reflow text within a rectangle), and
\texttt{glyph\_patch} (character-level correction). Micro-repair explicitly does not handle
geometry errors, missing placeholders, or containment violations---those failures still
trigger regeneration or replacement retry. After any edit, QA and figure insertion are rerun.
Figure~\ref{fig:micro_repair} shows an example.

\paragraph{Output.} The pipeline saves all intermediate artifacts under
\texttt{runs/<id>/}: \texttt{specs/} contains the full spec, storyboard, copy deck, and figure
selection; \texttt{prompts/} stores the complete prompt for every generation and critic call;
\texttt{generated/} holds critic-passing native templates and geometry-normalized versions;
\texttt{qa/} contains structured reports for placeholder QA, containment audit, and final QA;
\texttt{exports/} provides the final poster at native resolution and 4$\times$ upscaled, together
with a layered PowerPoint (\texttt{.pptx}) export in which the rendered layout is kept as a raster
slide background while each inserted source figure remains a separate, repositionable object rather
than being flattened into the image. The
\texttt{run\_manifest.yaml} records the full run status---every attempt, success or failure,
retry count, and failure category---so that any accepted final poster can be traced through
its complete generation history. A read-only observer additionally renders each run into a
standardized record---\texttt{run\_record.json} as a machine-readable source of truth and a
human-readable \texttt{run\_record.md}---that consolidates the objective provenance, aspect-contract,
containment, and public-text checks separately from the model-as-judge QA scores, so a run can be
audited without re-executing the pipeline.

\subsection{Contracts and Invariants}
\label{sec:contracts}

The harness is not merely a sequence of generation calls. It is organized around four
invariants that turn stochastic image generation into an auditable constrained workflow:

\paragraph{Placeholder contract invariant.} Every source scientific figure requested by
the poster specification must first appear as a labeled blank placeholder, not as a
model-drawn plot, diagram, table, microscopy panel, architecture figure, or spectrum. The
placeholder may contain only a figure ID, short public label, and aspect-ratio text. This
contract creates measurable instruction-following checks: placeholder count, label
legibility, blankness, and aspect-ratio compliance.

\paragraph{No-data-bearing-decoration invariant.} Decorative artwork may establish visual
style, domain atmosphere, or reading flow, but it must remain abstract. The model is not
allowed to add decorative fake axes, curves, Feynman-like graphs, molecule structures,
microscopy images, neural architecture diagrams, or other elements that a reader could
mistake for source-paper evidence.

\paragraph{Source insertion invariant.} Accepted figure regions are filled by compositing
source-paper assets into detected placeholder boxes. This invariant is narrower than full
poster factuality: it says that accepted figure regions are source-grounded, not that every
caption, text summary, or figure choice is automatically correct.

\paragraph{Strict failure propagation and audit invariant.} Geometry failures, missing
placeholders, containment violations, and public-text failures are explicit error states.
The system rejects or retries rather than silently hiding them.\footnote{Two narrow exceptions exist at the geometry level: a placeholder with no assigned source asset is skipped rather than raising (an unfilled region that the final figure-integration check then flags), and the purely decorative frame box is automatically widened to enclose its target. Neither inserts or fabricates figure content.} Each accepted artifact is
paired with prompts, structured specs, figure assignments, QA reports, and status manifests
so that failures and retries can be audited after generation.

\subsection{Domain-Profile Prompting}

\textsc{PosterHarness} includes a domain-profile prompting mechanism.
Rather than using a single, generic prompt template for all papers, the system can
identify the scientific domain and inject tailored design guidance. In this draft, this
mechanism is evaluated as extensibility and qualitative behavior; its quantitative gain
over generic prompting is left to the ablation protocol in Appendix~\ref{sec:ablation_plan}.

\paragraph{Domain classification.} The domain classifier sends a structurally sampled
excerpt (title, key sentences, figure captions) to the LLM, which assigns a primary domain
and provides a detection summary explaining the classification rationale. The classifier
is conservative: papers that cannot be reliably classified default to \texttt{generic},
which uses field-neutral scientific poster design grammar.

\paragraph{Domain profiles.} Each domain profile consists of three components:
\begin{itemize}
  \item \textbf{Visual grammar}: A domain-specific design spine (e.g., HEP: motivation
    → dataset/selection → strategy → result → interpretation; CS/ML: problem → method
    → experiments → quantitative results → limitations).
  \item \textbf{Figure composition rules}: Domain-specific guidance on figure hierarchy,
    placeholder sizing, and multi-panel treatment.
  \item \textbf{Text guidance}: Priority topics for domain-expert readers (e.g., HEP:
    luminosity, channel, SR/CR, systematic uncertainties; CS/ML: task, dataset, model,
    metrics, ablations).
  \item \textbf{Decorative art constraints}: Domain-specific prohibitions to prevent the
    model from generating fake versions of field-specific figures as decoration.
\end{itemize}

Profiles are defined declaratively in YAML configuration, making it straightforward
to add new domains without code changes. The narrative spines are manually specified
design defaults distilled from common paper structures in each field; they are not claims
that all HEP or CS/ML papers follow one universal order. The \texttt{auto} setting triggers
automatic classification; users can also manually specify a domain via
\texttt{--domain-profile}.

\paragraph{Content density presets.} Beyond visual grammar, domain profiles can
specify information density presets that control how much paper content survives into
the final poster. The framework provides two built-in configurations that operate
orthogonally to domain classification:
\begin{itemize}
  \item \texttt{standard} (default): Balanced information density suitable for most
    use cases. The paper content outline stage is disabled; the copy deck capacity is
    moderate (34 visible text units).
  \item \texttt{hep\_dense}: A higher-density preset motivated by HEP conference
    culture, where posters routinely carry detailed signal/control region definitions,
    systematic uncertainties, and fitted parameter values. This preset enables the
    content outline stage (up to 30 facts, 6 formulas), expands copy deck capacity
    to 42 units, and activates compact specialist text tiers.
\end{itemize}
The \texttt{hep\_dense} preset should be interpreted as one domain-specialized example
rather than a HEP-only feature. Researchers in other fields can add analogous presets---for
example, a biology preset emphasizing assay design and microscopy panels, a chemistry preset
emphasizing reaction schemes and spectra, or a CS/ML preset emphasizing architectures,
ablations, and benchmark tables. These presets are declarative prompt/profile extensions,
not changes to the replacement engine, and are selected independently of the domain profile.
This makes the framework adaptable to field-specific poster conventions as new communities
contribute their own prompt modules.

\subsection{Quality Assurance and Template Critic}

Quality assurance (QA) is integrated at every stage of the pipeline. The QA stages are
used as engineering gates and audit records; their model-judge scores are reported with
caveats and are not treated as ground truth:

\paragraph{Template critic.} After image generation, a VLM-based critic evaluates each
template across four numerically scored dimensions:
\begin{itemize}
  \item \textbf{Overall score} (threshold: 0.72): Holistic poster quality assessment.
  \item \textbf{Artistry score} (threshold: 0.65): Visual design, color harmony, typography.
  \item \textbf{Information density score} (threshold: 0.65): Adequacy of content placement
    and text legibility.
  \item \textbf{Placeholder contract score} (threshold: 0.75): Correctness of placeholder
    rendering (blank interior, correct labels, visible ratios).
\end{itemize}
Templates failing any dimension receive structured critique, and the generation is retried
with the critique appended to the prompt.

\paragraph{Placeholder QA.} After coordinate detection, the system verifies that each
placeholder region: (1) is blank and free of unintended content, (2) contains only the
specified ID, label, and aspect ratio text, (3) is unobstructed and properly padded.
A VLM-based QA stage performs visual inspection of the normalized layout.

\paragraph{Containment audit.} After replacement, the system deterministically verifies
that each pasted figure fits within its designated placeholder region.

\paragraph{Final QA.} The completed poster is inspected for public-text compliance (all
visible text must be public-facing with no internal workflow language), visual coherence,
and figure integration quality. A configurable list of forbidden phrases (e.g., ``TODO'',
``internal workflow'', ``placeholder explanation'') is checked.

\paragraph{Strict mode.} In strict mode, QA failures are explicit error states rather than
silent degradations. Failed templates are regenerated; failed replacements are retried with
adjusted parameters; and the run manifest records the rejection so that the failure remains
visible in later audits.

\subsection{Real Figure Replacement and Fidelity}

The deterministic replacement engine is the critical bridge for source-figure provenance.
It operates in three steps:

\begin{enumerate}
  \item \textbf{Box computation}: For each placeholder, four bounding boxes are computed
    from the detected coordinates---the target placement box, an erasure (cleanup) box,
    and the final visual frame box.
  \item \textbf{Containment verification}: Each target box is validated to lie strictly
    within its detected placeholder boundary. Violations raise hard errors.
  \item \textbf{Compositing}: Source figures are resized (preserving aspect ratio), pasted
    at target coordinates, and blended into the surrounding poster context with optional
    soft borders and shadows.
\end{enumerate}

This stage provides a narrower property than full scientific correctness: accepted
scientific figure regions are populated from source-paper assets rather than synthesized by
the image model. This does not by itself guarantee that all visible text, numerical summaries,
figure choices, or captions are correct; those require separate factuality checks and human or
VLM audit.

\section{Experimental Setup}

\subsection{Dataset and Paper Selection}

We construct a benchmark of \textbf{12 papers} across diverse scientific domains:

\begin{table}[!htbp]
\centering
\caption{Selected 12-paper benchmark used in this draft. HEP papers emphasize source-figure
provenance and specialist analysis content; non-HEP papers test transfer to AI/ML-adjacent
poster genres with different figure conventions.}
\label{tab:dataset}
\begin{tabular}{lcl}
\toprule
\textbf{Group} & \textbf{Count} & \textbf{Representative papers} \\
\midrule
High Energy Physics (HEP) & 6 & Higgs discovery, $H\to Z\gamma$, $W$ mass, WWZ/ZH \\
CS/ML and AI methods      & 5 & Transformer, BERT, NeRF, LoRA, GAT \\
Biology/ML                & 1 & Protein structure prediction \\
\midrule
\textbf{Total}            & \textbf{12} & \\
\bottomrule
\end{tabular}
\end{table}

Papers are selected to cover a range of visual complexity---from figure-heavy experimental
HEP analyses to architecture-heavy and result-heavy AI/ML papers. The 6 HEP papers include
CMS and ATLAS/CMS analyses with characteristic multi-panel limit plots, post-fit distributions,
control/validation-region figures, and collaboration-style result summaries. The 6 non-HEP
papers test whether the same placeholder-first mechanism transfers to transformer diagrams,
neural rendering examples, graph-learning schematics, low-rank adaptation plots, and protein
structure visualizations. Appendix~\ref{sec:paper_list} lists all papers and arXiv IDs.

\paragraph{Code and artifact release.} The full implementation, configuration files, and the
selected-12 benchmark artifacts are released at
\url{https://github.com/tyy99phy/paper_poster_harness}. The repository includes the complete
paired poster PNGs for all 12 papers, because reduced-size figures in a paper PDF are only
suitable as navigation previews rather than detailed visual inspection artifacts.

\paragraph{Cost-aware scope.} Full poster generation is not a negligible-cost benchmark: each
paper requires multiple LLM planning calls, at least one image-generation call, visual QA, and
sometimes regeneration. We therefore treat the current 12-paper set as a \textbf{pilot benchmark}
and a qualitative artifact release, not as a statistically exhaustive claim of dominance over
all scientific domains. If budget permits, the same protocol can be extended to 20 papers by
adding paired outputs rather than increasing the number of variants per paper.

\begin{table}[!htbp]
\centering
\caption{Cost-aware evaluation protocol. The expensive component is poster generation;
most measurements can be run repeatedly on already-generated PNGs and saved artifacts.}
\label{tab:cost_aware_protocol}
\small
\setlength{\tabcolsep}{4pt}
\begin{tabular}{p{0.22\textwidth}p{0.49\textwidth}p{0.19\textwidth}}
\toprule
\textbf{Evidence target} & \textbf{Low-cost measurement after posters exist} & \textbf{Claim type} \\
\midrule
Figure provenance & Source-asset provenance proxy and containment audit in this draft; small no-placeholder hallucinated-visual probe in Section~\ref{sec:prompt_ablation}; crop similarity reserved for future work & conditional pipeline claim \\
Information transfer & PosterQuiz-style and domain-fact questions answered from poster images & primary metric \\
Aesthetic preference & Blinded pairwise VLM ranking in this draft; preregistered human ranking reserved for future work & pilot metric \\
Robustness & Regeneration count, QA failures, final pass rate from manifests & system metric \\
Generalization & Add paired papers only until budget cap (12 $\to$ 20); keep baseline budget matched & scope claim \\
\bottomrule
\end{tabular}
\end{table}

\subsection{Baseline Systems}

We compare the following configurations:

\begin{itemize}
  \item \textbf{P2P / Paper2Poster}~\cite{pang2025paper2poster}: The state-of-the-art
    multi-agent poster generation system with PPTX-based rendering. We use the official
    pipeline and configure its internal agent backend to the same \texttt{gpt-5.5} model
    identifier used by our LLM/VLM stages, rather than a weaker fallback model.
  \item \textbf{PosterHarness} (standard): Our default pipeline with
    \texttt{domain\_profile: auto}, generic style, standard density preset,
    and all QA stages enabled. Structured planning, domain classification, template critic,
    QA, and micro-repair controller calls use \texttt{gpt-5.5}.
  \item \textbf{PosterHarness} (\texttt{hep\_dense}): The same pipeline with the
    HEP-specific high-density preset applied to 2 HEP papers, to evaluate whether
    configurable density benefits domain-expert audiences.
\end{itemize}

\paragraph{Model configuration and image-generation backend.} For controlled comparison,
both \textsc{PosterHarness} and Paper2Poster use \texttt{gpt-5.5} for LLM/VLM reasoning,
planning, critique, and text-processing stages. PosterHarness additionally uses OpenAI image
generation (GPT Image / ChatGPT Images 2.0 family, exposed in our deployment through the
\texttt{gpt-5.5} image-generation configuration)~\cite{openai2026imagegenerationdocs,openai2026chatgptimages2systemcard}
as the creative template designer. We
deliberately evaluate that backend only as a \emph{layout and typography engine}: all
scientific plots in accepted posters are later inserted from paper assets. This avoids
conflating OpenAI's improved text-rich visual generation with an unsupported claim that the
model can faithfully synthesize scientific data.

We also treat the image-generation interface as part of the experimental system rather than a
fully interchangeable implementation detail. OpenAI documents image generation through both a
direct Images API and the Responses API image-generation tool~\cite{openai2026imagegenerationdocs}.
The direct Images API provides a clean single-call endpoint and explicit output-size controls,
but its prompt field is documented as capped at 32{,}000 characters for GPT image
models~\cite{openai2026imagesapireference}. PosterHarness prompts can exceed this budget once
they include paper-specific content, domain grammar, placeholder geometry constraints, and
critic feedback; compressing the specification may reduce layout fidelity. We therefore use the
Responses-style image-generation route for the main long-context poster-template generation
path, while still auditing all generated candidates for canvas size, placeholder geometry, and
public-text compliance before deterministic figure insertion.

\subsection{Evaluation Axes}

We organize the evaluation around the research content of the harness rather than around
cosmetic scores alone. \emph{Source-figure provenance} asks whether accepted figure regions are
filled from paper assets rather than synthesized by the image model, with extraction,
assignment, and insertion errors treated as separate audit failures. \emph{Visual-summary
quality} asks whether the image-model template produces a different and richer design regime
than a deterministic PPTX-style baseline. \emph{Information transfer} asks whether the poster
still supports paper-specific question answering from the image alone. \emph{Harness behavior}
asks what failures are exposed by strict QA, how often retries occur, and what latency this
adds. We use simple visual proxies such as white-canvas fraction, color entropy, and native
resolution only to characterize the output regime; they are not substitutes for human or
expert judgments of poster quality.

This organization keeps the main claim focused: the contribution is the placeholder-first
architecture and its auditable workflow, not a claim that aesthetic proxies alone define
poster quality. The selected-12 benchmark provides evidence for the overall
system, while Section~\ref{sec:prompt_ablation} adds a counterfactual probe of the
placeholder contract itself. PosterQuiz-style scoring denotes an image-only information
retention test: questions derived from the source paper are answered from the rendered poster,
then normalized to a score in [0,1]. Human preference is deliberately deferred to a future
preregistered, externally recruited study rather than an informal in-house panel
(Appendix~\ref{sec:human_eval_protocol}). Table~\ref{tab:claim_evidence} summarizes
how each evaluation question is supported in this draft.

\begin{table}[!htbp]
\centering
\caption{Evaluation questions and evidence used in this draft. The table states the scope of
each result without turning pilot observations into broad generalization claims.}
\label{tab:claim_evidence}
\small
\setlength{\tabcolsep}{4pt}
\begin{tabular}{p{0.29\textwidth}p{0.43\textwidth}p{0.20\textwidth}}
\toprule
\textbf{Evaluation question} & \textbf{Evidence in this draft} & \textbf{Scope} \\
\midrule
Do accepted figure regions have source provenance? & Source-provenance proxy, corrected P2P figure cache, deterministic replacement records & Figure-region provenance, not full poster factuality \\
Does the visual regime differ from PPTX-like rendering? & Artifact statistics, high-resolution paired examples, blinded single-VLM preference & Pilot visual-summary comparison \\
What information-density trade-off appears? & PosterQuiz-style image-only scoring and answerability & Specialist completeness remains a target for dense modes \\
What does strict QA expose? & Runtime/status manifests, failure categories, and retry accounting & Harness transparency rather than per-module causality \\
How are scientific domains handled? & Declarative profile design and HEP/CS/ML/Bio examples & Extensibility mechanism with partial-domain evaluation \\
\bottomrule
\end{tabular}
\end{table}

\section{Pilot Benchmark Results}

The selected-12 benchmark is a cost-aware evaluation rather than a statistically
exhaustive one. It tests whether the proposed paradigm can simultaneously satisfy
three practical requirements: (i) no generated scientific plots in accepted figure regions,
(ii) visually rich poster-session artifacts that score well under a blinded single-VLM
audit, and (iii) enough paper-specific information to support reader understanding.

\subsection{Consolidated Quantitative Summary}

Table~\ref{tab:benchmark_summary} merges the previously separate artifact, provenance,
VLM-preference, PosterQuiz-style information retention, and runtime audits. This is the central quantitative table of
this draft: it presents a single benchmark view rather than scattering overlapping metrics
across multiple small tables.

\begin{table}[!htbp]
\centering
\caption{Cost-aware selected-12 benchmark summary. Blinded pairwise preference uses two
anonymized image orders per paper (24 comparisons total); one comparison is a score-derived
tie and is not assigned to either method. PosterQuiz-style scores are normalized to [0,1].}
\label{tab:benchmark_summary}
\small
\begin{tabular}{lcc}
\toprule
\textbf{Metric} & \textbf{Paper2Poster} & \textbf{PosterHarness} \\
\midrule
Accepted final PNGs & 12/12 & 12/12 \\
Source-figure provenance proxy & 12/12 & 12/12 \\
Native output size (Mpx) $\uparrow$ & $2.86\pm0.61$ & $25.17\pm0.00$ \\
White-canvas fraction $\downarrow$ & $0.677\pm0.035$ & $0.199\pm0.079$ \\
Color entropy (bits) $\uparrow$ & $2.15\pm0.22$ & $3.15\pm0.28$ \\
Edge/structure density $\uparrow$ & $0.122\pm0.013$ & $0.131\pm0.011$ \\
Blinded VLM score-derived wins & 3/24 & 20/24 \\
Blinded VLM overall score (1--5) & 3.38 & 4.42 \\
PosterQuiz-style information score & 0.940 & 0.918 \\
PosterQuiz-style answerable questions & 5.00/5 & 4.83/5 \\
Mean runtime (minutes) & $10.51\pm5.14$ & $27.96\pm17.95$ \\
\bottomrule
\end{tabular}
\end{table}

The results support a more nuanced claim than ``one system is better on every axis.''
\textsc{PosterHarness} receives higher single-VLM poster-session visual preference scores in this pilot
and produces a higher-resolution, less white-slide-dominated visual artifact. At the
same time, Paper2Poster remains slightly ahead on PosterQuiz-style information retention and is
faster, reflecting the advantage of a dense deterministic PPTX-style layout. This trade-off
is consistent with the design goal: our contribution is a new generation paradigm that
prioritizes source-figure provenance and editorial-quality poster design, not a claim
that every generated poster already maximizes factual density.

The released evaluation artifacts still include per-paper proxy plots, but we keep the main
paper focused on the consolidated table and the qualitative poster examples below.

\subsection{Run Audit and Failure Accounting}

The reviewer concern that a polished contact sheet can hide failed generations is important
for a system paper. We therefore report a run-level audit extracted from the benchmark status
manifests rather than only the accepted PNGs. During assembly of the selected benchmark,
\textsc{PosterHarness} produced 12 accepted final posters, but the status manifests record 22 selected-paper
job attempts: 12 successful runs, 8 failed or exception runs, and 2 cached/skipped attempts.
These failures were not silently folded into the accepted artifacts. The saved logs record
explicit strict-mode exits and failed jobs; observed examples include placeholder-geometry
rejection, placeholder-QA failure, transient backend errors, and final-QA rejection followed
by a fresh run. Table~\ref{tab:run_audit} summarizes the current audit.

\begin{table}[!htbp]
\centering
\caption{System run audit for the selected-12 benchmark. The table is computed from saved
status manifests and complements the accepted-poster metrics in Table~\ref{tab:benchmark_summary}.}
\label{tab:run_audit}
\small
\setlength{\tabcolsep}{5pt}
\begin{tabular}{lc}
\toprule
\textbf{Run-level quantity} & \textbf{Value} \\
\midrule
Accepted final \textsc{PosterHarness} posters & 12/12 \\
Recorded \textsc{PosterHarness} job attempts for selected papers & 22 \\
Successful attempts among recorded selected-paper jobs & 12 \\
Failed/exception attempts among recorded selected-paper jobs & 8 \\
Cached/skipped attempts & 2 \\
Failure-category audit rows & 8 \\
\textsc{PosterHarness} selected-run time (mean $\pm$ std; min--max) & $27.96\pm17.95$ min; 10.19--64.77 \\
Paper2Poster selected-run time (mean $\pm$ std; min--max) & $10.51\pm5.14$ min; 2.81--18.10 \\
Accepted source-figure provenance proxy & 12/12 \textsc{PosterHarness}; 12/12 corrected P2P cache \\
\bottomrule
\end{tabular}
\end{table}

This audit changes the interpretation of the system: the pipeline is not a one-shot generator
that always succeeds, but a strict harness that incurs additional latency to expose failures
and require retries before accepting a final poster. A stronger future study should compare
one-shot generation, critic-disabled generation, and the full harness under a fixed generation
budget; the recorded manifests and derived CSVs already make the cost of strict QA visible.

\subsection{Instruction-Following Failure Signals from Saved Logs}

The same manifests also make some instruction-following failures observable without running
new models. Table~\ref{tab:failure_categories} reports non-exclusive job-level categories
parsed from the eight failed or exception runs in the selected-paper assembly logs. These
counts should not be interpreted as rates for a fully controlled ablation: one failed job
can contain multiple rejection reasons, and some categories reflect infrastructure rather
than model behavior. They are nevertheless useful because they identify which contracts
actually blocked acceptance in the harness.

\begin{table}[!htbp]
\centering
\caption{Failure categories parsed from saved selected-paper logs. Counts are non-exclusive:
a failed job can contain more than one category. Source file:
\texttt{evaluation/posterharness\_failure\_category\_summary.csv}.}
\label{tab:failure_categories}
\small
\setlength{\tabcolsep}{4pt}
\begin{tabular}{lc}
\toprule
\textbf{Failure signal in saved logs/manifests} & \textbf{Failed/exception jobs with signal} \\
\midrule
Deterministic placeholder-geometry rejection & 7 \\
Strict placeholder-QA rejection & 6 \\
Template-critic rejection & 5 \\
Backend/account error & 1 \\
Public-text or notation issue mentioned & 1 \\
Launcher/file exception & 1 \\
\bottomrule
\end{tabular}
\end{table}

This table is an example of using scientific poster generation as a constrained
instruction-following audit. In the current run bundle, the most common recorded failures
are exactly the properties the placeholder-first contract is designed to expose:
aspect-ratio/geometry mismatch and placeholder cleanliness.

\subsection{Small No-Placeholder and Prompt-Component Probe}
\label{sec:prompt_ablation}

To close the most direct causal gap at small scale, we ran a scratch-only pilot probe on
three papers (two HEP papers and one CS/ML paper) without modifying the main harness
repository. For each paper we generated one image under four prompt conditions:
(i) \emph{direct no-placeholder}, which explicitly asks the image model to draw all scientific
figures itself; (ii) \emph{basic placeholders}, which asks only for blank [FIG NN] slots;
(iii) \emph{explicit placeholder contract}, which adds ID, blankness, aspect-ratio, and
no-data-bearing-decoration rules; and (iv) \emph{domain contract}, which adds the paper's
domain grammar to the explicit contract. A VLM audit then counted placeholder compliance and
scientific-visual violations. Table~\ref{tab:prompt_ablation} reports the aggregate results.
Counts are from one VLM auditor and are non-exclusive, so they should be treated as a pilot
instruction-following audit rather than a full benchmark.

\begin{table}[!htbp]
\centering
\caption{Small prompt-component probe on three papers. Ten figure slots were expected
across the three papers. Source file:
\texttt{evaluation/prompt\_component\_ablation\_summary.csv}.}
\label{tab:prompt_ablation}
\scriptsize
\setlength{\tabcolsep}{3pt}
\begin{tabular}{lcccccc}
\toprule
\textbf{Condition} & \textbf{Strict pass} & \textbf{IDs} & \textbf{Blank} &
\textbf{Aspect text} & \textbf{Synth. figs} & \textbf{Data-like decor.} \\
\midrule
Direct no-placeholder & 0/3 & 0/10 & 0/10 & 0/10 & 34 & 38 \\
Basic placeholders & 0/3 & 10/10 & 10/10 & 0/10 & 7 & 10 \\
Explicit placeholder contract & 0/3 & 10/10 & 10/10 & 10/10 & 0 & 4 \\
Domain contract & 1/3 & 10/10 & 10/10 & 10/10 & 2 & 10 \\
\bottomrule
\end{tabular}
\end{table}

\begin{figure}[!htbp]
\centering
\includegraphics[width=\textwidth]{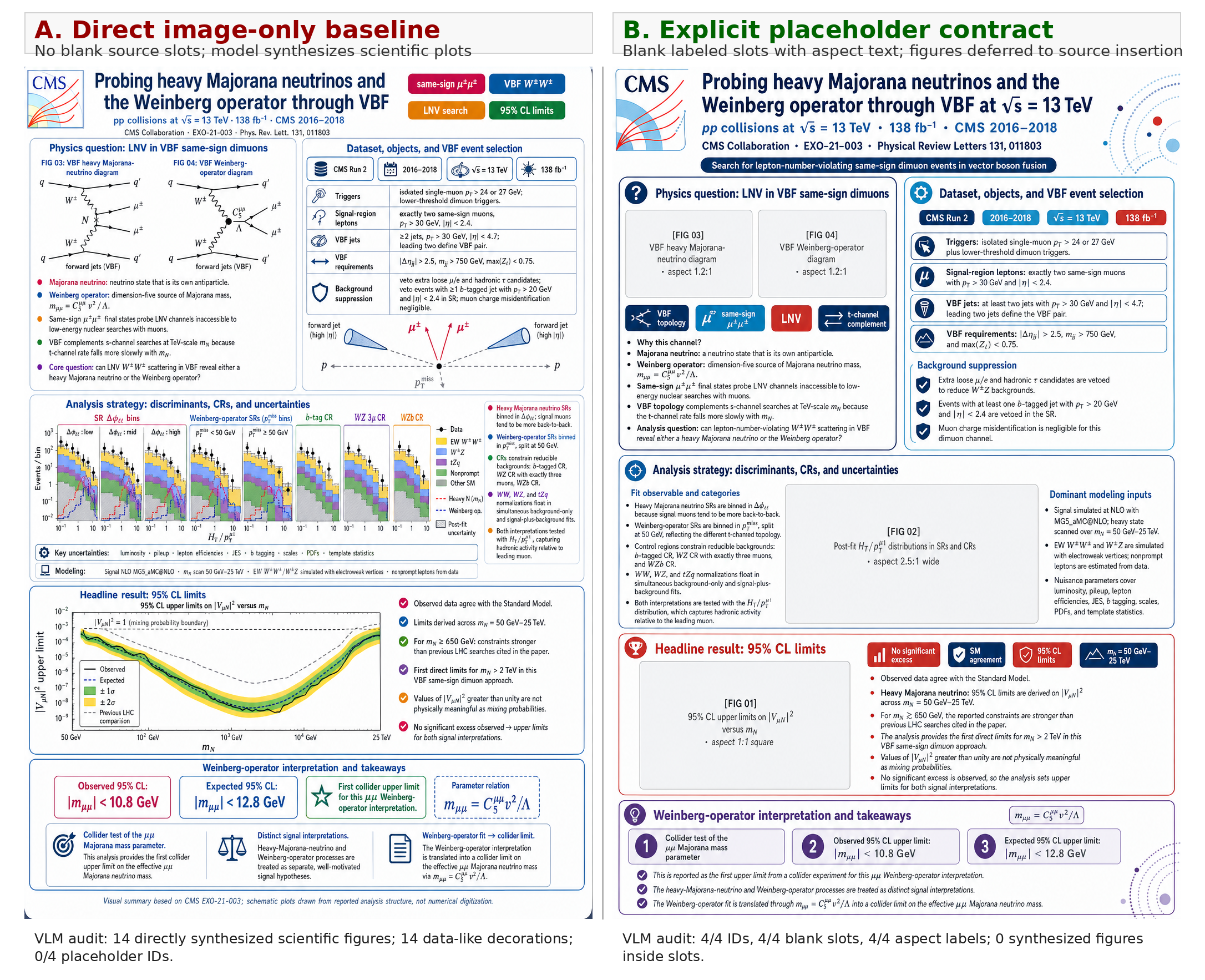}
\caption{Representative visual result from the no-placeholder probe on the SSWW HEP
paper. This figure isolates prompt compliance and is \emph{not} a final \textsc{PosterHarness}
output. Left: the direct image-only baseline produces visually plausible but model-synthesized
scientific plots and diagrams, leaving no auditable source slots. Right: the explicit
placeholder contract preserves blank labeled figure regions with aspect text; the full harness
would still need to run template QA, placeholder detection, deterministic source-figure insertion,
and final QA before acceptance.}
\label{fig:no_placeholder_ablation}
\end{figure}

The probe gives a concrete, evidence-backed version of the placeholder-first motivation.
When no placeholders are requested, the image model produces no auditable source slots and
instead synthesizes many scientific-looking figures. Basic placeholders recover the expected
slot IDs and blankness, but omit aspect-ratio text and still allow data-bearing decorations
outside the slots. The explicit placeholder contract fixes all observed slot IDs, blankness,
and aspect labels while reducing synthesized scientific content to zero in this small sample.
Adding domain grammar is not monotonic: it preserves slot compliance but can reintroduce
data-like domain decorations outside placeholders. This motivates keeping the no-data-bearing
decoration rule and visual QA as first-class contract checks rather than relying on domain
style prompts alone.

\subsection{Qualitative Results}

\begin{figure}[!htbp]
\centering
\includegraphics[width=\textwidth]{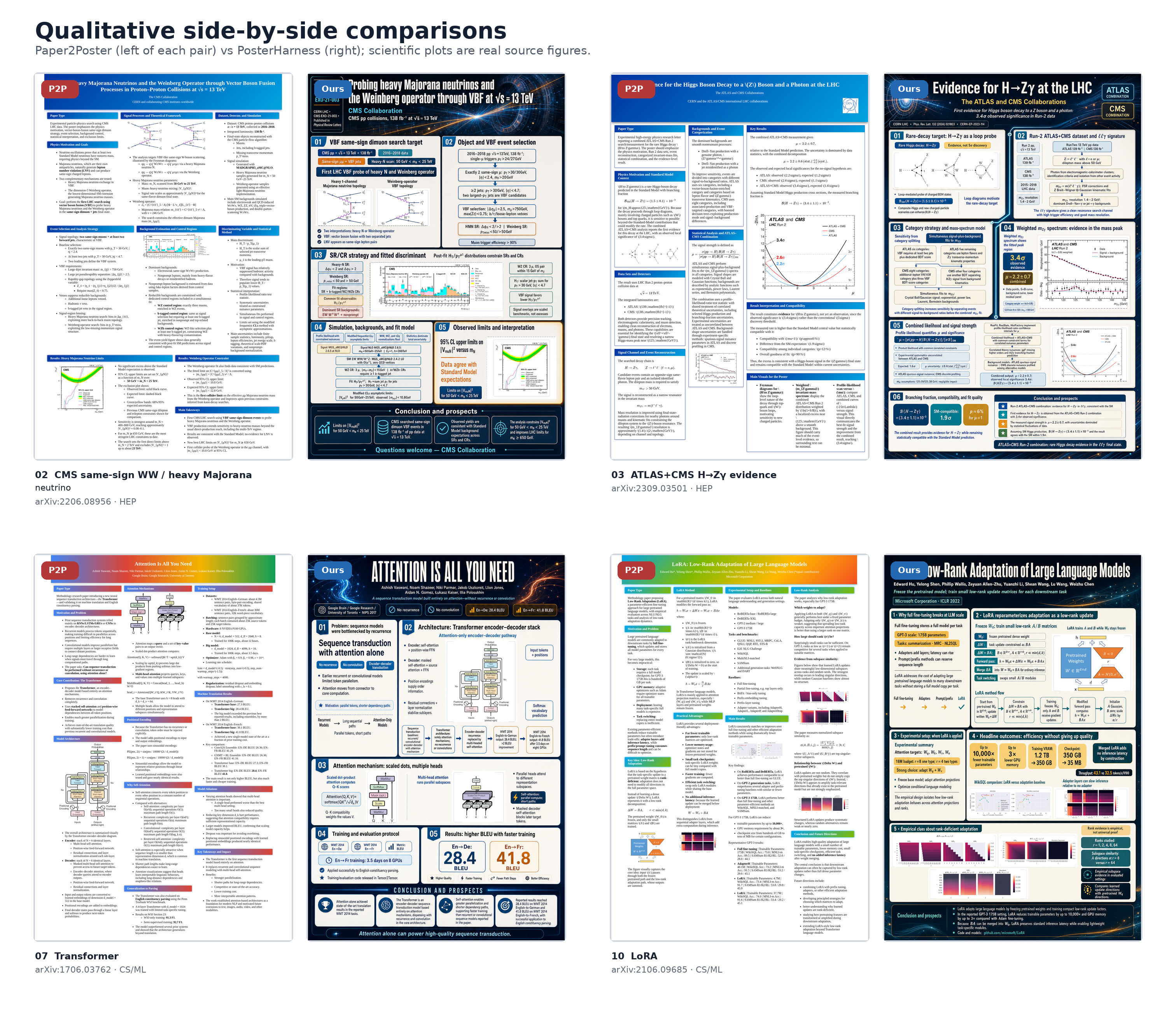}
\caption{Representative side-by-side comparisons from the selected benchmark.
Paper2Poster produces clean deterministic posters, but the visual language is often close to
a slide or PPTX grid. PosterHarness uses the image model for global composition, typography,
color, and background art, while preserving real scientific figures through deterministic
insertion. Full-resolution paired PNGs for all 12 papers are released in the repository.}
\label{fig:qualitative}
\end{figure}

\subsection{What the Benchmark Does and Does Not Prove}

The benchmark supports the paper's core engineering thesis: separating
creative template generation from source-figure insertion yields visually richer
posters while the image model is never asked to draw accepted scientific figure regions. It
does not establish universal superiority over deterministic poster renderers. PosterQuiz-style scoring
exposes the main trade-off---attractive layouts can omit dense quantitative details that a
specialist expects---which motivates the domain-specific dense modes discussed below.

\subsection{Threats to Validity}

Several choices in the current evaluation deliberately trade scale for artifact quality and
cost control. First, the 12 papers are curated examples rather than a random sample of
scientific literature; they support a pilot demonstration but not a population-level
generalization claim. Second, the VLM preference audit may favor visually rich artifacts over
more conservative but information-dense layouts; moreover, the single judge is drawn from the
same model family used to generate the templates, so a self-preference bias cannot be ruled
out. The preference result should therefore be read as single-judge pilot evidence and
interpreted together with PosterQuiz-style information scores and future multi-judge or human ratings. Third, our artifact proxies---white-canvas fraction,
color entropy, edge density, and native resolution---measure visual regime rather than
scientific usefulness by themselves. Fourth, Paper2Poster and PosterHarness optimize
different output formats: editable PPTX-style rendering versus a raster poster with stronger
editorial styling. Finally, all results depend on closed LLM/VLM and image-generation
backends whose behavior can change over time. These threats motivate the artifact release:
the repository records prompts, settings, paired PNGs, and low-cost evaluation scripts so
future work can rerun the same comparisons with additional models, ablations, and human
raters.

\section{Discussion}

\subsection{Key Insights}

\paragraph{The placeholder-first split narrows the strongest audit claim.} In a purely
image-generated poster, every plot-like region is a possible hallucination surface. In
\textsc{PosterHarness}, the strongest claim is more specific: accepted scientific figure
regions are filled from source-paper assets, conditional on correct extraction, assignment,
and placeholder detection. This source-provenance property is auditable from pipeline artifacts;
it should not be confused with a full guarantee of poster-level factuality.

\paragraph{Visual richness can be measured, not only shown.} The selected-12 artifacts show
large differences in native output resolution, white-canvas fraction, and global color
entropy. These metrics do not replace human judgment, but they provide objective evidence
that the outputs occupy a different design regime than the slide-like PPTX baseline.

\paragraph{Poster generation is a constrained instruction-following testbed.} The
placeholder contract makes failures measurable in ways that ordinary image-prompt examples
often do not: the evaluator can check placeholder count, placeholder ID accuracy,
blankness, target aspect ratio, scientific-decoration abstention, public-text hygiene, and
whether structured critique improves the next candidate. This reframes the harness as more
than an application demo: it is a way to study how text-rich image models follow scientific
layout contracts under audit.

\paragraph{Domain adaptation is a community extension point.} Qualitatively, HEP-specific
prompts and the \texttt{hep\_dense} preset produce more specialist content (SR/CR logic,
fit language, limits, and systematics). The more general lesson is that each scientific
community can encode its own expert expectations as declarative prompt presets. Future
benchmarks should quantify these presets with domain fact coverage and PosterQuiz-style scoring rather
than relying on visual examples alone.

\paragraph{Template QA should be evaluated as an engineering mechanism.} The critic and
placeholder-geometry checks make failures explicit and local: bad templates are regenerated
before source figures are inserted. The right quantitative measures are regeneration count,
pass rate, containment violations, and latency/cost, all of which can be extracted from run
manifests.

\paragraph{The no-placeholder probe supports the architectural split, but remains small.}
The scratch-only prompt probe in Section~\ref{sec:prompt_ablation} provides the first
direct counterfactual: when the same image-generation backend is asked to draw complete
scientific posters without placeholders, the VLM audit counts many synthesized scientific
figures and no auditable source slots. This supports the placeholder-first split, but only at
pilot scale. A stronger future study should repeat the comparison on a larger, budget-matched
set and add multiple VLM judges or blinded external raters.

\subsection{Toward Poster-as-Reading Interfaces}

A broader deployment direction is not limited to replacing a human-made conference poster.
A placeholder-first poster can act as a paper-reading interface: a reader opens an arXiv or
publisher page, requests a one-page visual summary, and receives a poster that combines an
editorial layout with real figures from the source document. This setting is especially
attractive for collaboration with scientific-search platforms, publishers, or model-assisted reading
systems because the system can expose a compact visual overview while preserving figure
provenance and intermediate QA artifacts. In such deployments, the poster need not be the
final editable artifact; it can be a trustworthy visual front-end for learning the main
claims, figures, and experimental setup of a paper.

\subsection{Limitations}

\begin{itemize}
  \item \textbf{Benchmark scale and coverage}: the current selected benchmark has 12 papers
    and is intentionally curated for clear comparison. Given image-generation cost, a near-term
    submission may realistically expand to 20 paired papers; broader 50--100 paper evaluation
    should be framed as future work rather than assumed evidence.
  \item \textbf{Human evaluation deliberately not reported}: this draft reports a single-VLM
    audit and PosterQuiz-style image-only scoring. Final user-utility claims require the
    externally recruited, blinded, preregistered study specified in
    Appendix~\ref{sec:human_eval_protocol}, with confidence intervals over papers and raters.
  \item \textbf{Output resolution}: the current image generation backend produces native
    1024×1536 templates, followed by high-resolution export/upscaling. Print-ready output
    ultimately requires vector/HTML/PDF reconstruction.
  \item \textbf{Editability}: the layered PPTX export now keeps each inserted source figure as a
    separate, repositionable object, but the surrounding generated layout---typography, reading
    path, and background art---is exported as a raster slide background, so full vector
    reconstruction would still be needed to make those elements editable as well.
  \item \textbf{Placeholder-count budget}: the current default caps selected source figures
    at eight. This is not a claim that eight figures are scientifically optimal; it is an
    empirical stability constraint. In local pilot runs, increasing the number of requested
    figure placeholders made the image model less reliable at following per-placeholder
    aspect-ratio instructions, especially for mixtures of wide, square, and portrait slots.
    Future versions should adaptively trade off figure count, absolute placeholder size, and
    multi-page or panelized poster layouts.
  \item \textbf{Image-generation interface dependence}: OpenAI documents both the direct
    Images API and the Responses API image-generation tool~\cite{openai2026imagegenerationdocs}.
    The direct Images API has clearer explicit output-parameter control, but its documented
    prompt budget for GPT image models~\cite{openai2026imagesapireference} can force compression
    of the long PosterHarness design specification. The Responses-style image-generation tool
    better matches the long-context template-generation path used here, but generated candidates
    remain subject to the same canvas-size, placeholder-geometry, and public-text audit gates.
  \item \textbf{Generation cost}: the pipeline uses premium image generation and repeated QA,
    so benchmark scale is directly constrained by compute/account budget. This makes careful
    paired evaluation and reusable low-cost audits essential.
  \item \textbf{Real-time generation}: the full pipeline requires minutes per poster on
    current infrastructure, limiting interactive use.
\end{itemize}

\section{Conclusion}

We presented \textsc{PosterHarness}, an auditable harness that turns scientific poster
generation into a measurable instruction-following benchmark for text-rich image models. The central idea is to give modern image generation the role it is
well suited for---global composition, typography, color, and editorial styling---while moving
data-bearing scientific figures into an auditable deterministic insertion path. This separation
turns scientific poster generation into a constrained visual-instruction-following problem:
figure regions must be blank, labeled, detectable, and replaceable by source-paper assets.

On the selected 12-paper benchmark---6 HEP analyses plus 6 non-HEP AI/ML-adjacent
papers---the artifacts demonstrate the core trade-off of the method. PosterHarness combines
image-model-driven visual design with deterministic source-figure provenance and receives
stronger single-VLM visual preference in this pilot, while Paper2Poster preserves slightly
more PosterQuiz-style information and runs faster. The run audit further shows that strict
QA is a visible engineering cost rather than a hidden detail: failed attempts and retries
are part of the harness behavior and should be reported with the final artifacts.

The small no-placeholder prompt probe strengthens the research claim behind the design:
when asked to draw complete scientific posters directly, the image model produces many
scientific-looking but synthesized visuals; explicit placeholder contracts turn those regions
into auditable slots. Future work should expand the benchmark beyond the cost-aware pilot set,
improve domain-specific dense modes, develop poster-as-reading interfaces with scientific
platforms, and reconstruct outputs as native vector graphics for print-quality production. We
release the pipeline, configuration, evaluation scripts, prompts, manifests, and benchmark
outputs as open-source resources for reproducible research in AI-assisted scientific
communication.

\bibliographystyle{unsrtnat}
\bibliography{references}

\newpage
\appendix

\section{Paper List}
\label{sec:paper_list}

\begin{table}[!htbp]
\centering
\caption{Complete selected benchmark paper list.}
\label{tab:paper_list}
\begin{tabular}{llll}
\toprule
\textbf{ID} & \textbf{Domain} & \textbf{arXiv} & \textbf{Short title} \\
\midrule
01 & HEP & 1207.7235 & CMS Higgs discovery \\
02 & HEP & 2206.08956 & CMS same-sign WW / heavy Majorana neutrino \\
03 & HEP & 2309.03501 & ATLAS+CMS H$\to$Z$\gamma$ evidence \\
04 & HEP & 2412.13872 & CMS W-boson mass \\
05 & HEP & 2409.13663 & CMS H$\to$ZZ$\to$4l mass and width \\
06 & HEP & 2505.20483 & CMS WWZ/ZH cross sections \\
07 & CS/ML & 1706.03762 & Transformer \\
08 & CS/ML & 1810.04805 & BERT \\
09 & CS/ML & 2003.08934 & NeRF \\
10 & CS/ML & 2106.09685 & LoRA \\
11 & CS/ML & 1710.10903 & Graph Attention Networks \\
12 & Bio/ML & 1911.05531 & Protein structure prediction \\
\bottomrule
\end{tabular}
\end{table}

\section{Repository Artifact Layout}
\label{sec:artifact_layout}

The public repository is \url{https://github.com/tyy99phy/paper_poster_harness}. We do
not include the complete selected-12 contact sheet in the paper PDF because the downsampled
version is not useful for detailed visual inspection. Instead, the repository serves as the
canonical artifact bundle. It contains:

\begin{itemize}
  \item \textbf{Framework code}: the poster generation CLI, domain profiles, prompt builders,
  placeholder QA, deterministic replacement, and evaluation utilities.
  \item \textbf{Selected benchmark posters}: for each selected paper, a pair of full-resolution
  PNGs, \texttt{ours.png} and \texttt{p2p.png}, organized by paper ID.
  \item \textbf{Evaluation outputs}: artifact metrics, runtime summaries, VLM pairwise judge
  results, PosterQuiz-style scoring outputs, run-attempt audit CSVs, and prompt packs for
  rerunning the audits.
  \item \textbf{Configuration records}: model and style settings used for the runs, including
  the \texttt{gpt-5.5} LLM/VLM configuration and the image-generation configuration used by
  \textsc{PosterHarness}.
\end{itemize}

This layout is intended to make the paper PDF a concise method-and-results document, while
leaving high-resolution visual inspection and rerunning of scripts to the versioned repository.

\section{Domain Profile Examples}
\label{sec:domain_profiles}

Domain profiles are declarative prompt modules rather than code branches. Each profile combines
visual grammar, figure composition rules, text priorities, and decorative constraints. Table~\ref{tab:domain_profiles}
shows representative examples.

\begin{table}[!htbp]
\centering
\caption{Example domain-profile components. Users can add analogous profiles or density presets
for their own disciplines without modifying the deterministic replacement engine.}
\label{tab:domain_profiles}
\scriptsize
\setlength{\tabcolsep}{3pt}
\begin{tabular}{p{0.14\textwidth}p{0.28\textwidth}p{0.25\textwidth}p{0.23\textwidth}}
\toprule
\textbf{Domain} & \textbf{Visual grammar} & \textbf{Content priorities} & \textbf{Figure treatment} \\
\midrule
HEP & Motivation $\rightarrow$ dataset/selection $\rightarrow$ analysis strategy $\rightarrow$ result $\rightarrow$ interpretation & Luminosity, collision energy, final state, SR/CR definitions, fit strategy, uncertainties, limits & Preserve official plots; avoid decorative fake histograms/Feynman graphs outside placeholders \\
CS/ML & Problem $\rightarrow$ architecture $\rightarrow$ training/evaluation $\rightarrow$ ablations $\rightarrow$ limitations & Task, dataset, model block, loss/objective, metrics, ablation results & Prioritize architecture diagrams and benchmark tables; allow schematic decorations only if non-data-like \\
Bio/ML & Biological question $\rightarrow$ data modality $\rightarrow$ model $\rightarrow$ qualitative evidence $\rightarrow$ quantitative result & Assay or dataset, sequence/structure modality, model objective, benchmarks, biological interpretation & Preserve microscopy/structure panels; avoid invented cells, spectra, or protein structures as decoration \\
Chemistry & Reaction/molecular motivation $\rightarrow$ method $\rightarrow$ characterization $\rightarrow$ yield/selectivity $\rightarrow$ mechanism & Reagents, conditions, spectra/characterization, yield, selectivity, mechanism hypothesis & Keep reaction schemes and spectra as source figures; decorative molecules must be abstract and non-claim-bearing \\
\bottomrule
\end{tabular}
\end{table}

The \texttt{hep\_dense} preset is one instance of this mechanism. A different community can
create a \texttt{bio\_dense}, \texttt{chem\_spectra}, or \texttt{ml\_benchmark} preset by
raising the copy budget and changing the domain-specific checklist used by the planner and
critic.

\section{Prompt Design Principles and Ablation Handles}
\label{sec:prompt_fragments}

The full prompts are long and generated from structured specifications, so we summarize the
most important reusable fragments here.

\paragraph{Placeholder contract.} The image model is instructed that every scientific figure
region must be a blank, clearly bounded placeholder containing only a figure ID such as
\texttt{[FIG 01]}, a short label, and the target aspect ratio. It must not draw curves,
axes, Feynman diagrams, microscopy images, architecture details, or any data-bearing content
inside those regions. Decorative art must stay abstract and outside scientific figure slots.

\paragraph{Domain-adaptive content planner.} The LLM receives the title, abstract, selected
paper sentences, figure captions, and asset manifest, then emits a schema-validated poster
specification: section titles, section roles, priority text units, selected source figures,
placeholder IDs, and a short reading-order storyboard. Domain profiles alter the checklist
used by this planner; for example, HEP profiles explicitly request final states, luminosity,
SR/CR definitions, and fit/result language.

\paragraph{Template critic.} The critic evaluates the generated template on four dimensions:
overall poster-session quality, artistry, information density, and placeholder-contract
compliance. A failing template returns structured critique that is appended to the next
generation attempt. Geometry or placeholder failures are not repaired by hiding them; they
trigger regeneration.

\paragraph{Micro-repair prompt.} Micro-repair prompts are intentionally narrow. They identify
a small public-facing defect, such as a localized title typo, and request an image edit that
fixes only that region while preserving the rest of the layout, placeholder boxes, and section
structure. They are not used for missing placeholders, wrong aspect ratios, or figure
containment failures.

Table~\ref{tab:prompt_components} summarizes the main prompt components as research
handles. The table is deliberately framed as an ablation interface, not as evidence that
each component has already been isolated experimentally.

\begin{table}[!htbp]
\centering
\caption{Prompt components, targeted failure modes, measurable audit metrics, and current
ablation status.}
\label{tab:prompt_components}
\scriptsize
\setlength{\tabcolsep}{3pt}
\begin{tabular}{p{0.18\textwidth}p{0.27\textwidth}p{0.28\textwidth}p{0.17\textwidth}}
\toprule
\textbf{Component} & \textbf{Targeted failure mode} & \textbf{Measurable audit metric} & \textbf{Current status} \\
\midrule
Placeholder contract & Model draws fake scientific plots or misses slots & Placeholder count/ID accuracy, blankness, aspect-ratio error & Enabled in main pipeline; isolated only in the small prompt-component probe \\
No-data-bearing decoration & Decorative art looks like source evidence & Scientific-decoration violation count outside placeholders & Small 3-paper prompt probe completed; larger multi-judge audit remains future work \\
Domain grammar & Generic content misses field-specific priorities & Domain fact coverage, PosterQuiz-style by domain, expert checklist coverage & Mechanism implemented; small probe shows non-monotonic decoration behavior, but domain-gain ablation not yet run \\
Layout/aspect labels & Wrong placeholder geometry breaks replacement & Deterministic placeholder-geometry rejection count & Logged; 7 failed/exception selected-paper jobs mention geometry rejection \\
Template critic feedback & Bad templates proceed or repeat the same error & Acceptance rate, retry count, failure category distribution & Logged in full pipeline; critic-disabled counterfactual not yet run \\
Micro-repair prompt & Local typo/header defect causes full rerun & Local edit success without changing geometry & Used only for local defects; not used for geometry/containment failures \\
\bottomrule
\end{tabular}
\end{table}

\section{Ablation and Failure Audit Status}
\label{sec:ablation_plan}

The current draft reports an end-to-end pilot benchmark plus the run audit in
Table~\ref{tab:run_audit}. To avoid over-claiming module-level causality, we separate
completed evidence from the ablations still required for a stronger systems paper.
Table~\ref{tab:ablation_plan} is therefore a status table rather than a list of hidden
positive results.

\begin{table}[!htbp]
\centering
\caption{Ablation and audit status. We report only completed audits as evidence; unrun
ablations are listed to delimit the present claims.}
\label{tab:ablation_plan}
\scriptsize
\setlength{\tabcolsep}{3pt}
\begin{tabular}{p{0.23\textwidth}p{0.40\textwidth}p{0.27\textwidth}}
\toprule
\textbf{Question} & \textbf{Ablation or statistic} & \textbf{Current status} \\
\midrule
Does placeholder-first matter? & Directly ask the image model to generate full scientific posters without placeholders; count hallucinated plots, fake axes, and fake scientific diagrams & Small 3-paper probe completed: direct no-placeholder produced 34 synthesized scientific figures counted by a VLM and 0/10 placeholder IDs; larger budget-matched ablation remains future work \\
Which prompt components matter? & Incrementally add placeholder contract, no-data-bearing-decoration rules, domain grammar, aspect labels, and critic feedback on a small fixed paper set & Small 3-paper probe completed for prompt components through domain grammar; critic-feedback counterfactual not yet run \\
Does domain adaptation help? & Auto/generic profile versus manually selected or domain-specific profile & Not yet run; current evidence is extensibility plus examples, not a quantified domain-gain claim \\
Does \texttt{hep\_dense} help specialists? & Standard density versus \texttt{hep\_dense} on HEP papers & Partially explored qualitatively; no formal fact-coverage ablation yet \\
Does the critic loop help? & Disable template critic or cap at one generation & Not yet run; current logs show retries/failures, but not the counterfactual without critic \\
Is VLM preference judge-dependent? & Repeat blinded A/B judging with several VLM judges and report agreement across visual appeal, skimability, trust, information density, and readability & Not yet run; current preference result is explicitly single-VLM pilot evidence \\
How reliable is replacement? & Report accepted source-figure provenance and final selected outputs & Completed at proxy level: 12/12 accepted \textsc{PosterHarness} outputs use deterministic real-figure insertion; full crop-level SSIM audit remains future work \\
What is the practical cost? & Log wall-clock time, selected-paper job attempts, and failure-category signals & Completed at manifest level: 22 recorded selected-paper job attempts, 12 successes, 8 failures/exceptions, 2 cached/skipped attempts; failure categories parsed from saved logs \\
\bottomrule
\end{tabular}
\end{table}

\section{Optional Human Evaluation Protocol}
\label{sec:human_eval_protocol}

We do not report informal human ratings because small unregistered panels can be biased by
familiarity with the authors or the method. A later version should instead use the following
externally recruited and blinded protocol.

\begin{enumerate}
  \item \textbf{Stimuli}: for each paper, show the Paper2Poster and PosterHarness outputs as
  anonymized A/B images. Randomize left/right order independently for each rater and paper.
  \item \textbf{Raters}: recruit a mixed panel of domain experts and non-expert scientific
  readers. HEP papers should include at least a small number of HEP-trained raters because
  SR/CR, fit, and limit information are specialist requirements.
  \item \textbf{Preference questions}: ask which poster is more likely to attract a viewer,
  which is easier to skim, which appears more scientifically trustworthy, and which the rater
  would prefer to use for a poster session.
  \item \textbf{Factual utility questions}: ask raters to answer short paper-specific questions
  from the poster alone, analogous to PosterQuiz-style scoring. For HEP, questions should cover process,
  luminosity/energy, event selection, fit/result, and supporting figures.
  \item \textbf{Analysis}: report win rate with bootstrap confidence intervals over papers,
  mean Likert scores per criterion, and answer accuracy for the factual questions. Ties and
  unreadable cases should be recorded explicitly rather than forced into binary preference.
\end{enumerate}

This protocol separates visual preference from information transfer, which is important
because the current pilot benchmark indicates a real trade-off: the single VLM audit favors
\textsc{PosterHarness} visually, while dense deterministic baselines can preserve slightly
more factual text.

\end{document}